\pdfoutput=1
\documentclass[11pt]{article}
\usepackage{geometry}
\usepackage{coling2020}
\usepackage{times}
\usepackage{url}
\usepackage{latexsym}
\usepackage{microtype}
\usepackage{booktabs}
\usepackage{multirow}
\usepackage{graphicx}
\usepackage{array}

\colingfinalcopy 

\title{LRG at SemEval-2020 Task 7:  Assessing the Ability of BERT and Derivative Models to Perform Short-Edits based Humor Grading}

\author{Siddhant Mahurkar\\
Department of Computer Science \& Engineering\\
Vellore Institute of Technology, Vellore, India\\
\tt siddhant.pravin2017@vitstudent.ac.in \AND
Rajaswa Patil\\
Department of Electrical \& Electronics Engineering\\
BITS Pilani K. K. Birla Goa Campus, India\\
\tt f20170334@goa.bits-pilani.ac.in}

\date{}

\begin{document}
\maketitle
\begin{abstract}
    In this paper, we assess the ability of BERT and its derivative models (RoBERTa, DistilBERT, and ALBERT) for short-edits based humor grading. We test these models for humor grading and classification tasks on the Humicroedit and the FunLines dataset. We perform extensive experiments with these models to test their language modeling and generalization abilities via zero-shot inference and cross-dataset inference based approaches. Further, we also inspect the role of self-attention layers in humor-grading by performing a qualitative analysis over the self-attention weights from the final layer of the trained BERT model. Our experiments show that all the pre-trained BERT derivative models show significant generalization capabilities for humor-grading related tasks.
\end{abstract}

\section{Introduction}
Humor is a communicative ability that produces a sense of laughter by imparting an amusing tone in conversations. If we can teach our models how to generate, detect, and grade humor in language, it would be a big step ahead in Natural Language Processing. Nearly all the existing datasets for humor-related tasks such as "Pun of the Day" \cite{yang2015humor} and "16000 One Liners" \cite{mihalcea2005making} focus on humor-detection in a single piece of text as a binary classification task. Previous tasks on humor-grading consist of datasets comprising of originally humorous texts \cite{potash-etal-2017-semeval,Chiruzzo2019OverviewOH}. SemEval-2020 Task-7 \cite{SemEval2020Task7} focuses on short-edits based humor-grading. The dataset used for this shared-task was introduced by \newcite{hossain-etal-2019-president}. It consists of short-edits applied to a piece of text (news-headlines), which changes it from non-funny to potentially funny. This shared-task consists of two independent sub-tasks based on this dataset. The first sub-task comprises of a regression task, where given the original and the edited news-headline, one must predict the funniness of the edited headline (humor-grading) on a scale of [0,3], where '0' conveys \emph{'least funny'} and '3' conveys \emph{'most funny'}. The second sub-task comprises of a classification task, where given two edited versions of the same news-headline, the model must predict the funnier one.

In this work, we propose using BERT \cite{devlin-etal-2019-bert} and its derivative models (collectively referred to as "BERT models" further throughout the paper) for short-edits based humor-grading. We mainly follow two approaches: In the first one we use pre-trained BERT models with a final regression layer to get a humor-grade for each edited news-headline, and for the second one we use masked-word language-modeling to fine-tune the pre-trained BERT models on the task dataset, which are further trained for the humor-grade regression task. We compare the results of these two approaches for each sub-task. We also test the generalization capabilities of the BERT models for this task by using the FunLines dataset \cite{hossain-etal-2020-funlines}, which is provided as an additional dataset for this shared-task. The code for this work is made publicly available as a GitHub repository.\footnote{\url{https://github.com/rajaswa/bert-humor-semeval-2020}}

\section{Background}
Early work in the field of humor detection was based on extracting various corpus-based statistical and linguistic features. \newcite{kiddon-brun-2011-thats} used these features with Support Vector Machine models for humor detection as a binary classification task. \newcite{zhang2014recognizing} worked on humor detection in tweets by extracting nearly fifty humor-related features. These features were used with Gradient Boosting Regression Tree based models. \newcite{inproceedings} showed that adding homophones and homographs as extra features to the existing linguistic features gave a small but significant improvement for humor detection in one-liner jokes. \newcite{chen2018humor} implemented a Convolutional Neural Network model with Highway Networks to train an end-to-end neural network for humor detection in English and Chinese languages.

More recent approaches for humor related tasks are focused around transformer-based architectures \cite{Ismailov2019HumorAB,mao2019bert,weller-seppi-2019-humor}. In this work, we experiment with BERT and its derivative transformer models. BERT \cite{devlin-etal-2019-bert} applies a bidirectional training of transformers with Masked Language Modelling and Next Sentence Prediction tasks. It is pre-trained on the Wikipedia (2,500 million words) and the Book Corpus (800 million words) datasets. RoBERTa \cite{DBLP:journals/corr/abs-1907-11692} is based on BERT but has been trained for a longer time with more amount of data (10 times that of BERT). Unlike BERT, RoBERTa is not trained for the Next Sentence Prediction task. DistilBERT \cite{Sanh2019DistilBERTAD} is a distilled version of BERT, with a 40\% size reduction over BERT while retaining 97\% of its language understanding capacity. ALBERT \cite{lan2019albert} is a lite BERT with an 89\% parameter reduction over the BERT-base model. It uses a self-supervised loss that focuses on inter-sentence coherence. BERT can be used with a classification head in humor detection tasks. A significant improvement over previous baselines proves that self-attention layers of BERT achieve success by extracting crucial humor-related features \cite{weller-seppi-2019-humor}. \newcite{Ismailov2019HumorAB} used a language-modeling based fine-tuning approach with BERT to get better results for humor-grading related tasks. These properties of BERT can be tested for generalization across the other BERT derivative models for short-edits based humor grading.

In this work, we test the ability of the BERT models to perform humor-grading tasks on the Humicroedit dataset (official task dataset) and FunLines dataset (additional dataset). Both of these are English short-edits based humor-grading datasets. The Humicroedit dataset consists of 15,095 edited headlines, and the FunLines dataset consists of 8,248 edited headlines. Both the datasets share the same format. Table~\ref{dataset-details} shows two different edits of the same news-headline with varying grades of humor.

\begin{table*}[t]
\centering
\renewcommand\arraystretch{2}
\begin{small}
\begin{tabular}[t]{>{\centering}m{0.3\linewidth}>{\centering}m{0.3\linewidth}>{\centering\arraybackslash}m{0.3\linewidth}}
\toprule
\textbf{Original Headline} & \textbf{Edited Headline} & \textbf{Humor Grade} \\
\midrule
US Navy ship fired \emph{\textbf{warning}} shots at an Iranian boat in the Persian Gulf & US Navy ship fired \emph{\textbf{tequila}} shots at an Iranian boat in the Persian Gulf & 2.6  \\
\midrule
US Navy ship fired \emph{\textbf{warning}} shots at an Iranian boat in the Persian Gulf & US Navy ship fired \emph{\textbf{texts}} shots at an Iranian boat in the Persian Gulf & 1.8  \\

\bottomrule
\end{tabular}
\end{small}
\caption{Example edited news-headlines from Humicroedit dataset}
\label{dataset-details}
\end{table*}

\section{Methodology}

\begin{figure}[t]
  \center
  \includegraphics[scale=0.85]{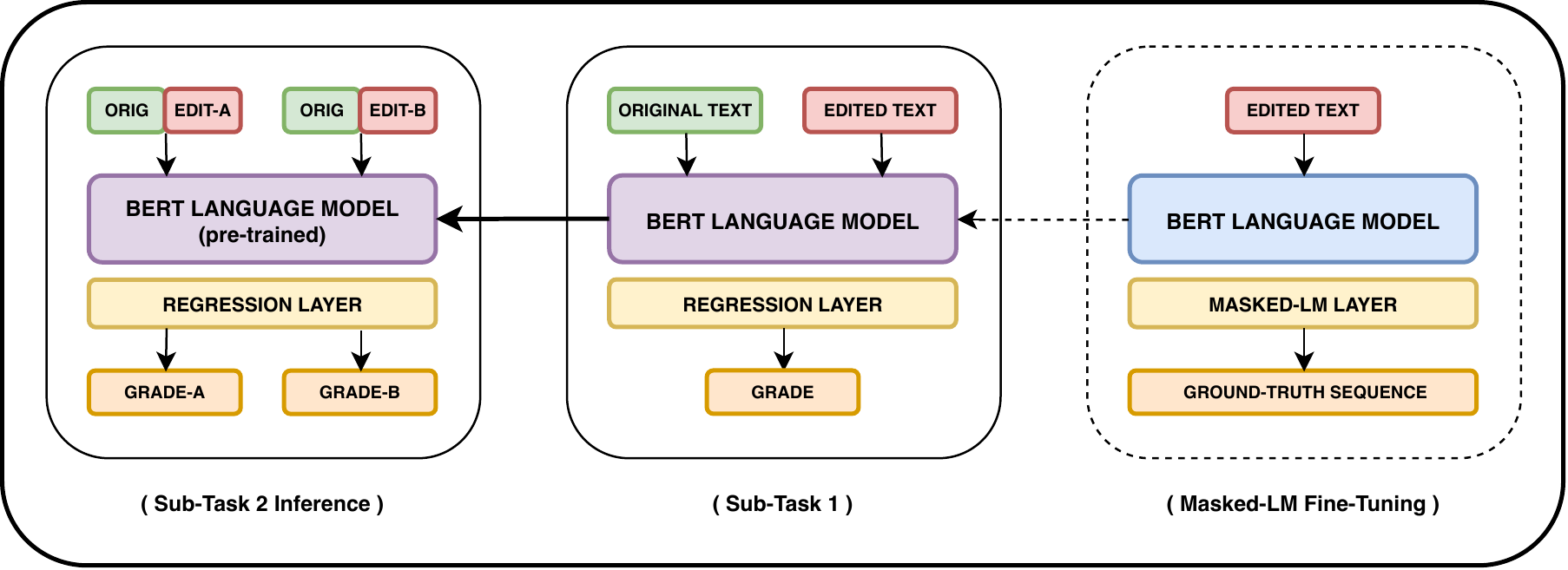}
  \caption{System Architecture}
  \label{architecture-fig}
\end{figure}

Since this is a short-edits based humor-grading dataset, the original text context is also crucial while grading the humor. To incorporate the original text context, we use a two-sentence input based approach with the BERT models, as shown in Figure~\ref{architecture-fig}. We concatenate the original and the edited news-headlines (separated and padded with special tokens of the respective BERT model) and feed it to the models in a tokenized format. Further, we inspect the effectiveness of masked language model fine-tuning based pre-training of the BERT models. Inspired from \newcite{Ismailov2019HumorAB}, we initialize the BERT models with their respective fine-tuned language models weights.

We follow a masked word prediction based language modeling approach to fine-tune the BERT language models on the entire dataset. Since the datasets do not come from an open-domain source and are limited to news headlines, all the words in the text are masked for prediction. We use only the edited humorous texts for language modeling pre-training. All the layers of the language models are trained with a maximum sequence length of 256 tokens for masked word prediction. The trained weights from these fine-tuned language models are used to initialize the model weights for sub-task 1. 

For sub-task 1, humor grading is considered a regression task over the humor grades of the edited news-headlines. We experiment with two types of weight-initializations for this task: weights from the language model fine-tuned on the dataset and the original pre-trained BERT model weights. We fine-tune the BERT models for the regression task by appending a fully-connected layer. This layer takes in the pooled output embedding from the BERT models and returns a humor grade value. The models are trained with the mean-squared error between the predicted and ground-truth humor grades as an objective function. The model weights are optimized for this objective function by Adam optimizer. For sub-task 2, we directly use the sub-task 1 models for a zero-shot inference over the pairs of edited news-headlines, as shown in Figure~\ref{architecture-fig}. After getting the humor grades for both the edits of a news-headline, we label the one with a higher grade value as the more humorous one. 

\section{Experimental Setup}
For all our experiments, we use the PyTorch\footnote{\url{https://pytorch.org/}} implementations of the smallest base variants of the pre-trained BERT models by Hugging Face\footnote{\url{https://huggingface.co/}} \cite{Wolf2019HuggingFacesTS}. The data distribution across the Training-Development-Test splits used for our experiments is shown in Table~\ref{data-split}. We validate our models with the mean-squared-error loss for the development set of sub-task 1. The official evaluation metric for sub-task 1 (humor grade regression) is the root-mean-squared-error (RMSE) loss. Whereas, the official evaluation metric for sub-task 2 (humor grade based multi-class classification) is the categorical accuracy. Since we directly use the sub-task 1 models for a zero-shot inference over the sub-task 2 data samples, there is no training involved for sub-task 2. Hence, we treat the entire sub-task 2 dataset (i.e., training + development + test) as the test set for our experiments. We also report the accuracy results on the official test set for the Humicroedit dataset. We create our own data splits for sub-task 1 on the FunLines dataset, as shown in Table~\ref{data-split}. Humicroedit and FunLines datasets share a similar format, but their backgrounds differ with respect to the annotations. To check the generalization of BERT models in terms of short-edits based humor detection, we perform cross-dataset inferences. For this,  we use models trained on the Humicroedit dataset for an inference over the FunLines dataset and those trained on the FunLines dataset for an inference over the Humicroedit dataset. Further, we use bert-viz tool by \newcite{vig2019transformervis} to get a visualization over the self-attention weights of the BERT models.

\begin{table}[h]
\centering
\begin{small}
\resizebox{0.6\textwidth}{!}{%
\renewcommand{\arraystretch}{0.01}
\begin{tabular}{@{}ccccc@{}}
\toprule
\textbf{Dataset} & \textbf{Sub-Task} & \textbf{Training} & \textbf{Development} & \textbf{Test} \\ \midrule
\multirow{2}{*}{Humicroedit} & 1 & 9652 & 2419 & 3024 \\ \cmidrule(l){2-5} 
 & 2 & 9381 & 2355 & 2960 \\ \midrule
\multirow{2}{*}{FunLines} & 1 & 5274 & 1322 & 1652 \\ \cmidrule(l){2-5} 
 & 2 & - & - & 1958 \\ \bottomrule
\end{tabular}%
}
\end{small}
\caption{Data distribution across the Training-Development-Test splits (number of samples)}
\label{data-split}
\end{table}

\section{Discussion}

For the final submission, we use BERT with masked word language model fine-tuning for both the sub-tasks. On the final test-set leaderboard, our system ranks 14\textsuperscript{th} for sub-task 1, with a RMSE loss of $0.53318$ and 13\textsuperscript{th} for sub-task 2, with an accuracy of $0.62177$. Further, in the post-evaluation experiments, we analyze the performance of all the BERT models on both the Humicroedit and the FunLines dataset. Table~\ref{results} shows the performance metrics of the BERT models both with and without the masked word language model fine-tuning along with the cross-dataset inferences. Overall, the BERT models perform better without masked word language model fine-tuning. This shows the capability of the BERT models to perform humor-grading tasks without the need for any language model fine-tuning on the humor dataset. All the BERT models perform significantly well for the cross-dataset inferences, showing their generalization capabilities across datasets from different backgrounds. Smaller BERT models (ALBERT: 11M parameters and DistilBERT: 65M parameters) do not show a significant drop in performance for sub-task 1 as compared to their bigger counterparts (BERT: 110M parameters and RoBERTa: 125M parameters). But we see a relatively lower accuracy on sub-task 2 for these smaller models. For sub-task 1, BERT slightly outperforms all the other models on both the datasets. Whereas for sub-task 2, RoBERTa gives a better performance than the rest of the models, with a significant gain on the FunLines dataset. Overall, all of the BERT models perform better on the Humicroedit dataset, which might be attributed to its relatively larger size as compared to the FunLines dataset.

\begin{table}[h]
\centering
\resizebox{\textwidth}{!}{%
\begin{tabular}{@{}cccccccc@{}}
\toprule
\multirow{2}{*}{\textbf{\begin{tabular}[c]{@{}c@{}}\\ BERT\\ DERIVATIVE\\ MODEL\end{tabular}}} & \multirow{2}{*}{\textbf{\begin{tabular}[c]{@{}c@{}}\\TRAINING\\DATASET\end{tabular}}} & \multirow{2}{*}{\textbf{\begin{tabular}[c]{@{}c@{}}\\MASKED-LM\\FINE\\TUNING\end{tabular}}} & \multicolumn{3}{c}{\textbf{HUMICROEDIT}} & \multicolumn{2}{c}{\textbf{FUNLINES}} \\ \cmidrule(l){4-8} 
 &  &  & \textbf{\begin{tabular}[c]{@{}c@{}}TASK-1 \\ (TEST SET)\\  (RMSE)\end{tabular}} & \textbf{\begin{tabular}[c]{@{}c@{}}TASK-2 \\ (TEST SET) \\ (ACC)\end{tabular}} & \textbf{\begin{tabular}[c]{@{}c@{}}TASK-2 \\ (ALL DATA) \\ (ACC)\end{tabular}} & \textbf{\begin{tabular}[c]{@{}c@{}}TASK-1 \\ (TEST SET)\\ (RMSE)\end{tabular}} & \textbf{\begin{tabular}[c]{@{}c@{}}TASK-2 \\ (ALL DATA) \\ (ACC)\end{tabular}} \\ \midrule
\multirow{4}{*}{\begin{tabular}[c]{@{}c@{}}\\\textbf{BERT}\\ (base-uncased)\end{tabular}} & \multirow{2}{*}{Humicroedit} & Yes & 0.533\textsuperscript{*} & 0.622\textsuperscript{*} & 0.622 & 0.643 & 0.605 \\ \cmidrule(l){3-8} 
 &  & No & \textbf{0.53} & 0.619 & \textbf{0.643} & 0.601 & 0.616 \\ \cmidrule(l){2-8} 
 & \multirow{2}{*}{FunLines} & Yes & 0.671 & 0.601 & 0.598 & 0.527 & 0.6 \\ \cmidrule(l){3-8} 
 &  & No & 0.631 & 0.614 & 0.604 & \textbf{0.521} & 0.677 \\ \midrule
\multirow{4}{*}{\begin{tabular}[c]{@{}c@{}}\\\textbf{RoBERTa}\\ (base)\end{tabular}} & \multirow{2}{*}{Humicroedit} & Yes & 0.533 & \textbf{0.628} & 0.641 & 0.591 & 0.628 \\ \cmidrule(l){3-8} 
 &  & No & 0.541 & 0.596 & 0.643 & 0.591 & 0.596 \\ \cmidrule(l){2-8} 
 & \multirow{2}{*}{FunLines} & Yes & 0.668 & 0.507 & 0.499 & 0.56 & 0.493 \\ \cmidrule(l){3-8} 
 &  & No & 0.696 & 0.587 & 0.581 & 0.541 & \textbf{0.718} \\ \midrule
\multirow{4}{*}{\begin{tabular}[c]{@{}c@{}}\\\textbf{ALBERT}\\ (base-v2)\end{tabular}} & \multirow{2}{*}{Humicroedit} & Yes & 0.582 & 0.537 & 0.552 & 0.669 & 0.516 \\ \cmidrule(l){3-8} 
 &  & No & 0.577 & 0.55 & 0.556 & 0.658 & 0.548 \\ \cmidrule(l){2-8} 
 & \multirow{2}{*}{FunLines} & Yes & 0.673 & 0.521 & 0.524 & 0.572 & 0.553 \\ \cmidrule(l){3-8} 
 &  & No & 0.704 & 0.517 & 0.525 & 0.568 & 0.541 \\ \midrule
\multirow{4}{*}{\begin{tabular}[c]{@{}c@{}}\\\textbf{DistilBERT}\\ (base-uncased)\end{tabular}} & \multirow{2}{*}{Humicroedit} & Yes & 0.572 & 0.541 & 0.536 & 0.647 & 0.541 \\ \cmidrule(l){3-8} 
 &  & No & 0.562 & 0.577 & 0.574 & 0.648 & 0.572 \\ \cmidrule(l){2-8} 
 & \multirow{2}{*}{FunLines} & Yes & 0.678 & 0.543 & 0.541 & 0.557 & 0.536 \\ \cmidrule(l){3-8} 
 &  & No & 0.67 & 0.545 & 0.549 & 0.551 & 0.563 \\ \bottomrule
\end{tabular}%
}
\caption{Post Evaluation Experiments (*final submission)}
\label{results}
\end{table}

\begin{figure}[h]
    \center
  \includegraphics[scale=0.9]{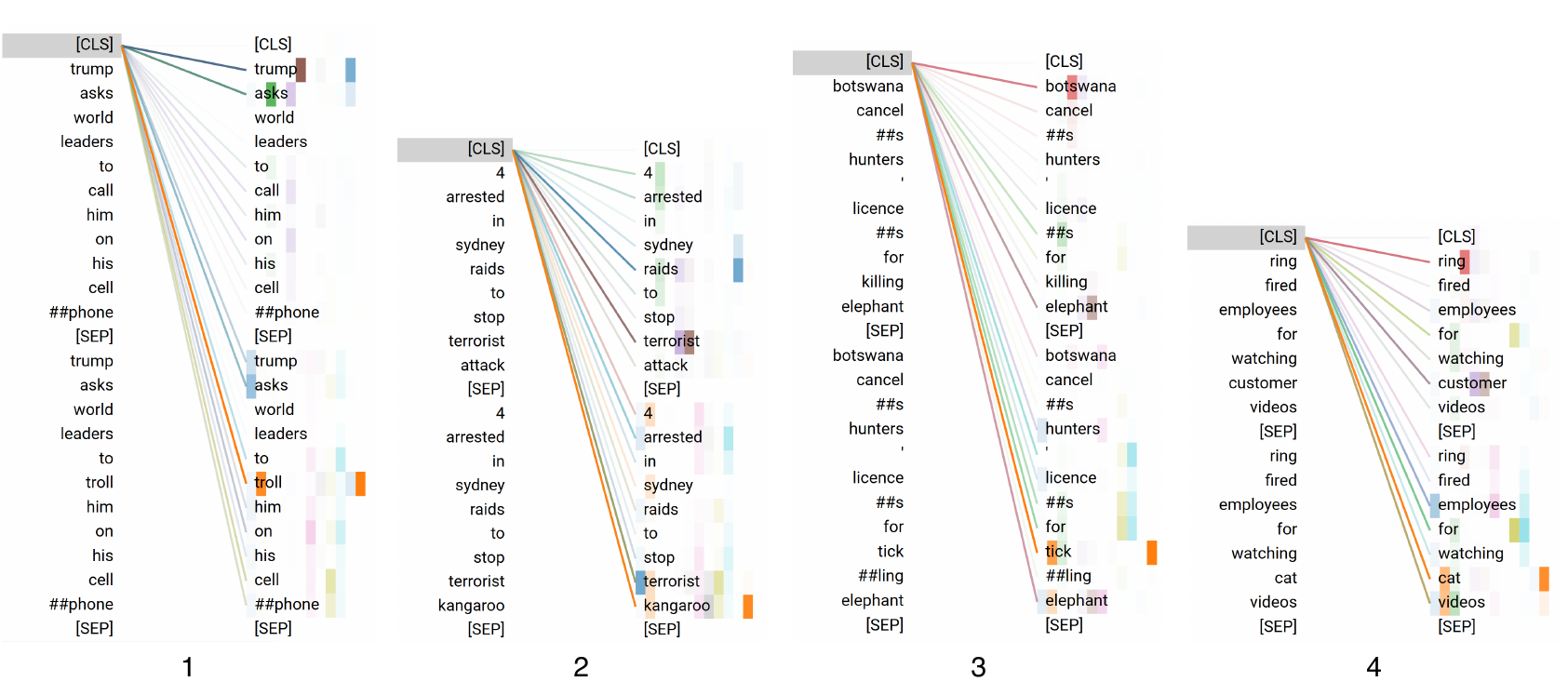}
  \caption{Attention Visualisation over self-attention weights of BERT model}
  \label{bert-visualization}
\end{figure}

BERT models use the [CLS] token as the pooled output representation for the entire input sequence. The [CLS] token is ultimately used for the humor-grading tasks in our experiments. In order to interpret the BERT models' decisions, we analyze the multi-head self-attention weights of the [CLS] token from the final layer of the trained BERT model, as shown in Figure~\ref{bert-visualization}. Taking a closer look at the attention-visualization over these attention weights reveals that the model learns to assign more attention to certain parts of the text, which play a role in imparting funniness to the text as a whole. In general, some attention heads (2\textsuperscript{nd} and 12\textsuperscript{th} head) learn to assign more attention to the edited words as compared to other words in the text. The attention from the 2\textsuperscript{nd} and 12\textsuperscript{th} head is represented by orange-colored connections in Figure~\ref{bert-visualization}.

Figure~\ref{bert-visualization}-1 and Figure~\ref{bert-visualization}-3 represent the attention visualization of some of the best predictions from the Humicroedit and FunLines dataset, respectively. In these cases, we observe that the [CLS] token firmly attends the edited words by the 2\textsuperscript{nd} and 12\textsuperscript{th} attention head. On the other hand, Figure~\ref{bert-visualization}-2 and Figure~\ref{bert-visualization}-4 represent some of the worst predictions from the Humicroedit and FunLines dataset, respectively. Here, the [CLS] token divides its attention relatively more uniformly across the entire input sequence, with relatively lesser attention from the 2\textsuperscript{nd} and 12\textsuperscript{th} attention head. The same trend is observed across most of the samples from both of the datasets for all the BERT models. This reveals the importance of self-attention in the decision making of the BERT models for short-edits based humor grading.

\section{Conclusion}
In this work, we tested the ability of BERT and its derivative models for short-edits based humor-grading tasks. Our experiments showed that the BERT models perform better with their original pre-trained weights, and there is no need for language model fine-tuning on the humorous texts. The tests also revealed that the BERT models display good generalization capabilities for humor-grading tasks by performing significantly well with zero-shot inference across the sub-tasks and cross-dataset inferences. The qualitative analysis over the self-attention weights of the BERT model revealed the attentive bias shown by some of the attention heads towards the edited parts of the text. It also showed the importance of self-attention in the decision making ability of the BERT models.

We plan to extend our work by testing autoregressive and generative transformer models for humor-grading related tasks. One can also experiment with different language modeling techniques and different input formats for the short-edits based humor-grading tasks. To further test the abilities of the BERT models for humor-grading, we plan to test our approach on other humor-related datasets such as the "Pun of the Day" and the" 16000 One Liners" datasets.

\bibliographystyle{coling}
\bibliography{semeval2020}

\begin{thebibliography}{}

\bibitem[\protect\citename{Beukel and Aroyo}2018]{inproceedings}
Sven Beukel and Lora Aroyo.
\newblock 2018.
\newblock Homonym detection for humor recognition in short text.
\newblock 10.

\bibitem[\protect\citename{Chen and Soo}2018]{chen2018humor}
Peng-Yu Chen and Von-Wun Soo.
\newblock 2018.
\newblock Humor recognition using deep learning.
\newblock In {\em Proceedings of the 2018 Conference of the North American
  Chapter of the Association for Computational Linguistics: Human Language
  Technologies, Volume 2 (Short Papers)}, pages 113--117.

\bibitem[\protect\citename{Chiruzzo \bgroup et al.\egroup
  }2019]{Chiruzzo2019OverviewOH}
Luis Chiruzzo, Santiago Castro, Math{\'i}as Etcheverry, Diego Garat,
  Juan~Jos{\'e} Prada, and Aiala Ros{\'a}.
\newblock 2019.
\newblock Overview of haha at iberlef 2019: Humor analysis based on human
  annotation.
\newblock In {\em IberLEF@SEPLN}.

\bibitem[\protect\citename{Devlin \bgroup et al.\egroup
  }2019]{devlin-etal-2019-bert}
Jacob Devlin, Ming-Wei Chang, Kenton Lee, and Kristina Toutanova.
\newblock 2019.
\newblock {BERT}: Pre-training of deep bidirectional transformers for language
  understanding.
\newblock In {\em Proceedings of the 2019 Conference of the North {A}merican
  Chapter of the Association for Computational Linguistics: Human Language
  Technologies, Volume 1 (Long and Short Papers)}, pages 4171--4186,
  Minneapolis, Minnesota, June. Association for Computational Linguistics.

\bibitem[\protect\citename{Hossain \bgroup et al.\egroup
  }2019]{hossain-etal-2019-president}
Nabil Hossain, John Krumm, and Michael Gamon.
\newblock 2019.
\newblock {``}president vows to cut {\textless}taxes{\textgreater} hair{''}:
  Dataset and analysis of creative text editing for humorous headlines.
\newblock In {\em Proceedings of the 2019 Conference of the North {A}merican
  Chapter of the Association for Computational Linguistics: Human Language
  Technologies, Volume 1 (Long and Short Papers)}, pages 133--142, Minneapolis,
  Minnesota, June. Association for Computational Linguistics.

\bibitem[\protect\citename{Hossain \bgroup et al.\egroup
  }2020a]{SemEval2020Task7}
Nabil Hossain, John Krumm, Michael Gamon, and Henry Kautz.
\newblock 2020a.
\newblock Semeval-2020 {T}ask 7: {A}ssessing humor in edited news headlines.
\newblock In {\em Proceedings of International Workshop on Semantic Evaluation
  (SemEval-2020)}, Barcelona, Spain.

\bibitem[\protect\citename{Hossain \bgroup et al.\egroup
  }2020b]{hossain-etal-2020-funlines}
Nabil Hossain, John Krumm, Tanvir Sajed, and Henry Kautz.
\newblock 2020b.
\newblock {S}timulating creativity with funlines: A case study of humor
  generation in headlines.
\newblock In {\em Proceedings of {ACL} 2020, System Demonstrations}, Seattle,
  Washington, July. Association for Computational Linguistics.

\bibitem[\protect\citename{Ismailov}2019]{Ismailov2019HumorAB}
Adilzhan Ismailov.
\newblock 2019.
\newblock Humor analysis based on human annotation challenge at iberlef 2019:
  First-place solution.
\newblock In {\em IberLEF@SEPLN}.

\bibitem[\protect\citename{Kiddon and Brun}2011]{kiddon-brun-2011-thats}
Chlo{\'e} Kiddon and Yuriy Brun.
\newblock 2011.
\newblock That{'}s what she said: Double entendre identification.
\newblock In {\em Proceedings of the 49th Annual Meeting of the Association for
  Computational Linguistics: Human Language Technologies}, pages 89--94,
  Portland, Oregon, USA, June. Association for Computational Linguistics.

\bibitem[\protect\citename{Lan \bgroup et al.\egroup }2019]{lan2019albert}
Zhenzhong Lan, Mingda Chen, Sebastian Goodman, Kevin Gimpel, Piyush Sharma, and
  Radu Soricut.
\newblock 2019.
\newblock Albert: A lite bert for self-supervised learning of language
  representations.
\newblock {\em arXiv preprint arXiv:1909.11942}.

\bibitem[\protect\citename{Liu \bgroup et al.\egroup
  }2019]{DBLP:journals/corr/abs-1907-11692}
Yinhan Liu, Myle Ott, Naman Goyal, Jingfei Du, Mandar Joshi, Danqi Chen, Omer
  Levy, Mike Lewis, Luke Zettlemoyer, and Veselin Stoyanov.
\newblock 2019.
\newblock Roberta: {A} robustly optimized {BERT} pretraining approach.
\newblock {\em CoRR}, abs/1907.11692.

\bibitem[\protect\citename{Mao and Liu}2019]{mao2019bert}
Jihang Mao and Wanli Liu.
\newblock 2019.
\newblock A bert-based approach for automatic humor detection and scoring.

\bibitem[\protect\citename{Mihalcea and Strapparava}2005]{mihalcea2005making}
Rada Mihalcea and Carlo Strapparava.
\newblock 2005.
\newblock Making computers laugh: Investigations in automatic humor
  recognition.
\newblock In {\em Proceedings of the Conference on Human Language Technology
  and Empirical Methods in Natural Language Processing}, pages 531--538.
  Association for Computational Linguistics.

\bibitem[\protect\citename{Potash \bgroup et al.\egroup
  }2017]{potash-etal-2017-semeval}
Peter Potash, Alexey Romanov, and Anna Rumshisky.
\newblock 2017.
\newblock {S}em{E}val-2017 task 6: {\#}{H}ashtag{W}ars: Learning a sense of
  humor.
\newblock In {\em Proceedings of the 11th International Workshop on Semantic
  Evaluation ({S}em{E}val-2017)}, pages 49--57, Vancouver, Canada, August.
  Association for Computational Linguistics.

\bibitem[\protect\citename{Sanh \bgroup et al.\egroup
  }2019]{Sanh2019DistilBERTAD}
Victor Sanh, Lysandre Debut, Julien Chaumond, and Thomas Wolf.
\newblock 2019.
\newblock Distilbert, a distilled version of bert: smaller, faster, cheaper and
  lighter.
\newblock {\em ArXiv}, abs/1910.01108.

\bibitem[\protect\citename{Vig}2019]{vig2019transformervis}
Jesse Vig.
\newblock 2019.
\newblock A multiscale visualization of attention in the transformer model.
\newblock {\em arXiv preprint arXiv:1906.05714}.

\bibitem[\protect\citename{Weller and Seppi}2019]{weller-seppi-2019-humor}
Orion Weller and Kevin Seppi.
\newblock 2019.
\newblock Humor detection: A transformer gets the last laugh.
\newblock In {\em Proceedings of the 2019 Conference on Empirical Methods in
  Natural Language Processing and the 9th International Joint Conference on
  Natural Language Processing (EMNLP-IJCNLP)}, pages 3621--3625, Hong Kong,
  China, November. Association for Computational Linguistics.

\bibitem[\protect\citename{Wolf \bgroup et al.\egroup
  }2019]{Wolf2019HuggingFacesTS}
Thomas Wolf, Lysandre Debut, Victor Sanh, Julien Chaumond, Clement Delangue,
  Anthony Moi, Pierric Cistac, Tim Rault, R'emi Louf, Morgan Funtowicz, and
  Jamie Brew.
\newblock 2019.
\newblock Huggingface's transformers: State-of-the-art natural language
  processing.
\newblock {\em ArXiv}, abs/1910.03771.

\bibitem[\protect\citename{Yang \bgroup et al.\egroup }2015]{yang2015humor}
Diyi Yang, Alon Lavie, Chris Dyer, and Eduard Hovy.
\newblock 2015.
\newblock Humor recognition and humor anchor extraction.
\newblock In {\em Proceedings of the 2015 Conference on Empirical Methods in
  Natural Language Processing}, pages 2367--2376.

\bibitem[\protect\citename{Zhang and Liu}2014]{zhang2014recognizing}
Renxian Zhang and Naishi Liu.
\newblock 2014.
\newblock Recognizing humor on twitter.
\newblock In {\em Proceedings of the 23rd ACM International Conference on
  Conference on Information and Knowledge Management}, pages 889--898.

\end{thebibliography}

\end{document}